\documentclass[10pt,letterpaper]{article}
\usepackage[top=0.85in,left=2.75in,footskip=0.75in,marginparwidth=2in]{geometry}

\usepackage[utf8]{inputenc}

\usepackage{cite}

\usepackage{nameref,hyperref}

\usepackage[right]{lineno}
\usepackage{amsmath,amsfonts}

\usepackage{microtype}
\DisableLigatures[f]{encoding = *, family = * }

\raggedright
\usepackage{changepage}

\usepackage[aboveskip=1pt,labelfont=bf,labelsep=period,singlelinecheck=off]{caption}

\makeatletter
\renewcommand{\@biblabel}[1]{\quad#1.}
\makeatother

\usepackage{lastpage,fancyhdr,graphicx}
\usepackage{epstopdf}
\fancyheadoffset[L]{2.25in}
\fancyfootoffset[L]{2.25in}

\usepackage{color}

\definecolor{Gray}{gray}{.25}

\usepackage{graphicx}

\usepackage{sidecap}

\usepackage{wrapfig}
\usepackage[pscoord]{eso-pic}
\usepackage[fulladjust]{marginnote}
\reversemarginpar

\begin{document}
\vspace*{0.35in}

\begin{flushleft}
{\Large
\textbf\newline{Teaching reproducible research for medical students and postgraduate pharmaceutical scientists.}
}
\newline
\\
Andreas D. Meid\textsuperscript{1,*}
\\
\bigskip
\bf{1} Department of Clinical Pharmacology and Pharmacoepidemiology, University of Heidelberg, Im Neuenheimer Feld 410, 69120 Heidelberg, Germany
\\
\bigskip
* andreas.meid@med.uni-heidelberg.de 

\end{flushleft}

\section*{Abstract}
In many academic settings, medical students start their scientific work already during their studies. Like at our institution, they often work in interdisciplinary teams with more or less experienced (postgraduate) researchers of pharmaceutical sciences, natural sciences in general, or biostatistics. All of them should be taught good research practices as an integral part of their education, especially in terms of statistical analysis. This includes reproducibility as a central aspect of modern research. Acknowledging that even educators might be unfamiliar with necessary aspects of a perfectly reproducible workflow, I agreed to give a lecture series on reproducible research (RR) for medical students and postgraduate pharmacists involved in several areas of clinical research. Thus, I designed a piloting lecture series to highlight definitions of RR, reasons for RR, potential merits of RR, and ways to work accordingly. In trying to actually reproduce a published analysis, I encountered several practical obstacles. In this article, I focus on this working example to emphasize the manifold facets of RR, to provide possible explanations and solutions, and argue that harmonized curricula for (quantitative) clinical researchers should include RR principles. I therefore hope that these experiences are helpful to raise awareness among educators and students. RR working habits are not only beneficial for ourselves or our students, but also for other researchers within an institution, for scientific partners, for the scientific community, and eventually for the public profiting from research findings.

\linenumbers

\section*{Background and aims}
I recently gave a lecture series on reproducible research (RR) for medical students and postgraduate pharmacists, because we considered reproducibility to be of topical importance for them. During the course, I made useful experiences that I would like to share hoping to encourage reproducible working habits, to emphasize the crucial role of supervisors, and thus strengthen our awareness how important RR is when teaching good research practices.

\section*{General experiences}
During the course, I approached RR following the key questions about what it is by definition, why we should work accordingly, by which means we can conduct reproducible analyses, and how we can profit from them. The very first finding from the course was indeed that different researchers use the terms "reproducible" and "replicable" differently and sometimes interchangeably. Following the "new lexicon for research reproducibility"\textsuperscript{1}, (methods) reproducibility is based on the same research conditions and is a fundamental requirement for successful replication (results reproduction). Full replication instead involves independent investigators, independent data, and optionally independent methods.\textsuperscript{2} \\
On this basis, we explored notorious negative examples (e.g., the "Duke-scandal" leading event termination of clinical trials and lawsuits after article retraction\textsuperscript{3}) and worked out advantages for the single researcher (e.g., streamlined working habits, strengthened confidence, easier adaptations), the research team (e.g., higher research impact), and the (scientific) community (e.g., better public and inter-professional recognition). We introduced ourselves to several concepts and technical solutions (e.g., literate programming with R-markdown), and realized the advantages of common standards. Likewise, working examples made us aware of practical challenges, too.

\section*{A working example}
As an illustrative hands-on example in the course, I picked a publication heading into a current direction with high prospects for personalized medicine, namely predicting individual benefit by exploring heterogeneous treatment effects (HTE) in the SPRINT and ACCORD trials (\href{https://www.clinicaltrials.gov}{https://www.clinicaltrials.gov}: unique identifiers NCT01206062 and NCT00000620).\textsuperscript{4} Based on the available baseline information, the authors developed models predicting individual treatment response to intensive antihypertensive treatment so that (better) treatment decisions could be made. Interestingly, a machine learning approach called \textit{X}-learner outperformed several alternative methods including the conventional logistic regression with interaction terms. The publication cites a publicly available repository where the project is shared and is generally a very positive example bypassing several known barriers to proper reproducibility, among them poor standardization of model building, lacking or insufficient documentation, incomplete transparency, or coding and typing errors. 
Inspired by this favourable setting, I decided to reproduce the published findings. Initially, ethical committee approval for data availability had to be obtained before data access could be granted. After that, a workspace environment for the \textit{Python} programming language had to be set up. This was not straightforward to me, but an intuitive \textit{readme} file was as helpful as other elements related to ‘best practice’ ideas, e.g., standardized folders and structure or so-called \textit{make} files (synonymous with \textit{run}-scripts or \textit{do}-files) for automatic data loading and running of the analysis code. The repository in particular provides several conceptual benefits when it comes to making distinctions between source data and derived files. It also helped to recognize the dependencies between code elements, different files, or libraries. In the end, the project could indeed be run and yielded results in the identical fashion to the published paper. 
The published analysis apparently satisfied all three levels of reproducibility, full \textit{depth} (i.e., code for data preparation, analytical results, and figures), \textit{portability} to another computer system, and full \textit{coverage} (i.e., all published results). Nevertheless, slightly diverging results and the fact that Python module versions have changed since the original publication made me curious to translate the \textit{Python} code to \textit{R} code as a more customary solution in medical research. This would also mimic the situation of any colleague being not familiar with dedicated analysis software. Interestingly, resulting estimates from my attempt to reproduce analyses were even more intriguing. \textbf{Figure 1} illustrates several metrics from the original publication describing the performance of the X-learner in predicting individual treatment benefits to intensified blood pressure control.
\begin{adjustwidth}{-2in}{0in}
\begin{center}
\includegraphics[width=115mm]{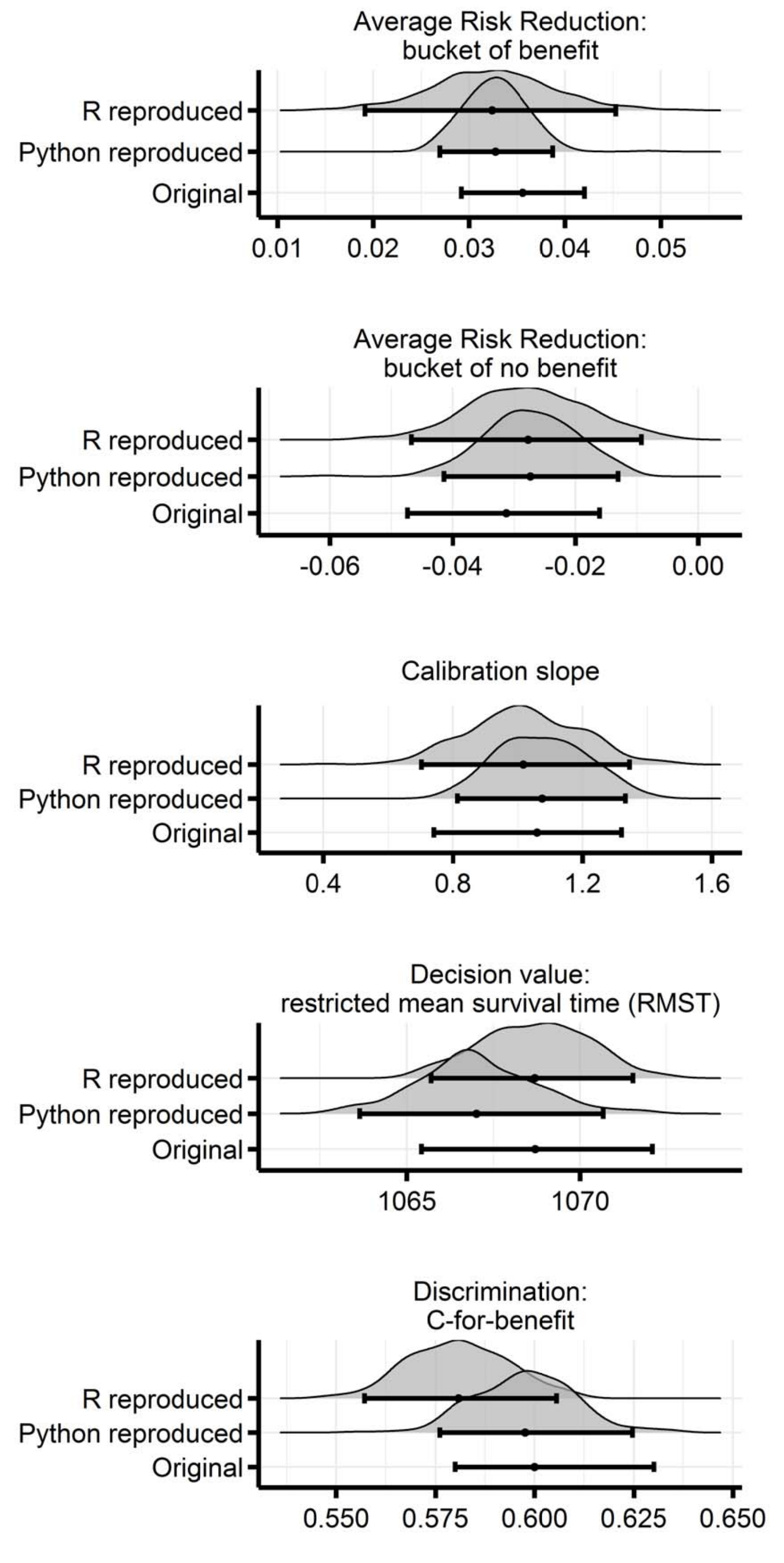}
\end{center}
\justify 
\color{Gray}
\textbf {Figure 1. Bootstrapped performance metrics used to derive mean estimates and 95 \% confidence intervals from the original publication\textsuperscript{4}, from the reproduction using the supplied code written in Python, and from the supplied code translated to R.}
Average risk reductions were calculated for two subgroups (buckets) of those patients with predicted benefit in absolute risk reduction (ARR $>$ 0) and those patients without predicted benefit (ARR $\leq$ 0). A calibration line was fitted between quintiles of ARRs and predicted risk, whose slope is chosen for this set of performance metrics. As a decision value, the model predicted restricted mean survival time (RMST [\textit{days}]) indicates the mean time to event if treatment choice would have been based on the predicted individual benefit (and is thus to be compared with the baseline value of 1061.24 days, 95 \% confidence interval: [1057.37; 1064.10]). The \textit{concordance-statistic for benefit} is a metric reflecting the model’s ability to predict treatment benefit (rather than risk for an outcome). Using the \textit{Python} implementation to calculate this metric, the individual risk estimates reproduced in R yielded an estimate of 0.61 [0.55; 0.72]. Of note, we restrict the presentation of results to distributions from resampling and their summary parameters; further numerical metrics to quantify reproducibility are left out for simplicity. Analyses were using the Anaconda distribution of \textit{Python} version 3.7.3 (Anaconda Software Distribution, version 2-2.4.0) and the \textit{R} software environment version 3.6.1 (R Foundation for Statistical Computing, Vienna, Austria).
\end{adjustwidth}
\vspace{.5cm} 
All performance metrics relied on bootstrapped samples to derive estimates for their mean and 95 \% confidence intervals. Among several possible explanations, the fact that random forests are indeed based on a random process is the most conclusive argument, especially when they have to be reproduced with different programs on different operating systems (and thus different \textit{seeds}).\textsuperscript{5}

\section*{Findings and conclusions}

These experiences from the lecture series’ working example expose several noteworthy aspects. Good documentation, standardized workflows, available data, and a freely available software solution facilitate RR. In this particular case, this framework likewise helped to teach and better understand cutting-edge methods. By deciding to reproduce this excellent work, I tried to provide a prototypical positive example and simultaneously elucidated barriers that any of us could have trying to reproduce these findings. Considering the potential benefits of RR and our surprising observations and obstacles, the key question is how to incorporate these insights into typical workflows of medical research that could guide educating supervisors and students. 
Hands-on education and scientific interaction stand on top of all requirements. In addition to publications, tutorials, or courses, the principles of RR should also appear in harmonized curricula for quantitative medical researchers. While a clear roadmap for instructors would have yet to defined, practical teaching is generally advisable: our hands-on experiences from a fully reproduced analysis illustrated good practices and practical obstacles for achieving strictly identical results. Scientific interactions might include discussion forums or workshops being held either face-to-face or virtually. The encouraging experiences from my pilot lecture project strongly suggest an interdisciplinary mixture of participants to fully profit from different perspectives and from a different educational background. If these efforts further supported reproducible projects, it would not only build confidence and credibility within our broad discipline, but also towards other disciplines and decision-makers. In order for our findings to impact regulatory decisions or patient care, results must be replicable in independent settings as the ultimate standard, for which reproducible projects are fundamental.\textsuperscript{2}



\nolinenumbers

\section*{Funding}
The author is funded by the Physician-Scientist Programme of Heidelberg University, Faculty of Medicine. The funder has no role in any actions related to this analysis, or preparation of the paper. Thus, the author declares no conflicts of interest and is responsible for the content and writing of the article alone.

\section*{Conflicts of interest}
The author declares that he has no conflict of interest. 

\section*{Availability of data and material}
This manuscript reports the experiences from a lecture series about reproducible research. For teaching purposes, a published analysis of particular trial data was reproduced, for which ethics committee approval (reference number V-223/2019 at the University of Heidelberg) and permission from the data keeping institution BioLINCC (\href{https://biolincc.nhlbi.nih.gov/home/}{https://biolincc.nhlbi.nih.gov/home/}) were obtained. Access to the underlying data of the published analysis\textsuperscript{4} being reproduced for teaching purposes can be requested at \href{https://biolincc.nhlbi.nih.gov/home/}{https://biolincc.nhlbi.nih.gov/home/}. As specified in published analysis\textsuperscript{4}, the corresponding analysis code is available from the repository \href{https://github.com/tonyduan/hte-prediction-rcts}{https://github.com/tonyduan/hte-prediction-rcts}.

\bibliography{library.bib}

\bibliographystyle{abbrv}

\end{document}